\pdfoutput=1

\documentclass[11pt]{article}

\usepackage[preprint]{acl}

\usepackage{times}
\usepackage{latexsym}
\usepackage{amssymb}
\usepackage{pifont}
\newcommand{\xmark}{\ding{55}}%

\usepackage[T1]{fontenc}

\usepackage[utf8]{inputenc}

\usepackage{microtype}

\usepackage{inconsolata}

\usepackage{amsmath}
\usepackage{times}
\usepackage{latexsym}
\usepackage{booktabs}
\usepackage{microtype}
\usepackage[pdftex]{graphicx}
\usepackage{subfigure}
\usepackage{color,soul} 
\usepackage{makecell}
\usepackage{subcaption}
\usepackage{framed} 
\title{Evaluating LLMs for Quotation Attribution in Literary Texts: A Case Study of LLaMa3}

\begin{NoHyper}

\author{Gaspard Michel$^{\dagger\ast}$ \\ \texttt{gmichel@deezer.com} \And Elena V. Epure$^\dagger$  \\ \texttt{eepure@deezer.com}\And
  Romain Hennequin$^\dagger$ \\ \texttt{rhennequin@deezer.com} \AND Christophe Cerisara$^\ast$ \\ \texttt{christophe.cerisara@loria.fr} \AND
  $^\dagger$ \normalfont{Deezer Research, Paris, France} \\ $^\ast$ \normalfont{Loria, Nancy, France}
 }
\end{NoHyper}

\begin{document}
\maketitle
\begin{abstract}
Large Language Models (LLMs) have shown promising results in a variety of literary tasks, often using complex memorized details of narration and fictional characters. 
In this work, we evaluate the ability of Llama-3 at attributing utterances of direct-speech to their speaker in novels. 
The LLM shows impressive results on a corpus of 28 novels, surpassing published results with ChatGPT and encoder-based baselines by a large margin.
We then validate these results by assessing the impact of book memorization and annotation contamination.
We found that these types of memorization do not explain the large performance gain, making Llama-3 the new state-of-the-art for quotation attribution in English literature. We release publicly our code and data\footnote{\url{https://github.com/deezer/llms_quotation_attribution}}.
\end{abstract}

\section{Introduction}

Quotation attribution, or the automated attribution of utterances to fictional characters, is of crucial importance for character analysis in digital humanities \cite{Elson2010, Muzny2017a, Labatut2019, Sims2020}.
However, quotation attribution remains a challenging task, and recent approaches still struggle to find methods that generalize across writing styles.
A few works have explored the use of LLMs for quotation attribution in novels, by extracting conversations directly with ChatGPT \cite{Zhao2024} or by asking ChatGPT to attribute a single quote given its surrounding context \cite{Su2023}.
Yet, these works do not propose a systematic evaluation of LLMs for quotation attribution in literary works. 


Another significant evaluation drawback in assessing LLMs is the lack of analysis regarding book memorization and annotation contamination, which can hinder their generalization abilities.
Book memorization occurs when an LLM is able to generate specific passages of texts in a novel, and is correlated with its frequency in pretraining data \cite{Carlini2023}.
In contrast, data contamination arises when an LLM has memorized evaluation data, enabling it to produce labels without reasoning \cite{magar-schwartz-2022-data}.
To avoid confusion, we refer to data contamination as annotation contamination. Addressing both issues is essential when evaluating LLMs on literary tasks, as they can significantly impact the understanding of its performance on downstream tasks.


In this work, we start by evaluating the performance of Llama-3 8b on the Project Dialogism Novel Corpus (PDNC) \cite{Vishnubhotla2022}, a corpus of 28 English novels.
We selected Llama-3 8b due to its popularity, its impressive performance on various tasks \cite{dubey2024llama}, and because its pretraining corpus only includes data up to March 2023, which makes the second release of PDNC annotations not included in the pretraining data.
We carefully designed prompts with Chain-of-Thought reasoning \cite{wei2022chain}, and use the larger context size of LLMs to directly attribute all quotes in a given chapter.
Our results indicate that this method improves attribution accuracy compared to predicting a single quote in a contextual passage. 
We next conduct an evaluation of book memorization and annotation contamination to determine whether Llama-3's success stems from its reasoning abilities or its capacity to memorize passages and annotations.

We found that our Llama-3 based approach demonstrates remarkable performance, improving attribution accuracy by 12 points against state-of-the-art systems on the first 22 novels on PDNC and by 9 points on the remaining novels.
Besides, we could not find signs of annotation contamination on the first 22 PDNC novels, and we show that although memorization impacts speaker predictions on a subset of quotes, a majority of successful predictions can be attributed to the reasoning ability of Llama-3.
We validate this finding by evaluating the LLM on a recently published novel not included in its pretraining data, where our approach performs on-par with the current state-of-the-art system, BookNLP+ \cite{Vishnubhotla2023, michel-etal-2024-improving}. Besides, we found that our approach combined with the larger Llama-3 70b reaches an almost perfect accuracy.
To sum up, our contributions are: 


\begin{enumerate}
    \item We evaluate Llama-3 zero-shot performance on PDNC, comparing it to strong systems and show a major accuracy improvement on PDNC novels, establishing a new state-of-the-art for quotation attribution accuracy on English literature.
    \item We introduce a novel measure of book memorization, \textit{Corrupted-Speaker-Guessing}, that classifies a successful quote attribution into either a reasoning or memorization prediction.
    We propose this new measure as other metrics \cite{chang2023} failed to detect memorization of canonical literature when used with Llama-3 8b. 
    We validate our measure following a similar evaluation protocol as  \citet{chang2023}.
    \item We thoroughly evaluate the impact of book memorization and annotation contamination on the downstream task, showing that these memorization types are not the principal factors of Llama-3 quotation attribution accuracy.
    
\end{enumerate}


\section{Related Work}

\paragraph{LLMs for literary tasks} Large Language Models (LLMs) have shown promising results in a variety of literary tasks related to Narrative Understanding \cite{Xu2023, Underwood2023, piper-bagga-2024-using, hobson-etal-2024-story, bamman2024classificationlargelanguagemodels} or Character Understanding and Profiling \cite{Soni2023, Yu2023a}.
Their capacity of memorizing important details of fictional characters has also been studied for character understanding \cite{Stammbach2022, Zhao2024, Wang2024}.
In this work, we assess LLMs on the quotation attribution task systematically by accounting for memorization and annotation contamination. For this, we introduce a new measure of book memorization and show that Llama-3's state-of-the-art results are not explained by memorization but rather by its reasoning ability.

\paragraph{Quotation Attribution}
Methods to attribute direct speech to its speaker in literary texts have explored sequence labeling \cite{okeefe-etal-2012-sequence}, deterministic rules \cite{Muzny2017} or generation \cite{Su2023}.
BookNLP, a popular Natural Language Processing pipeline dedicated to books, also proposes a quotation attribution system that was recently improved \cite{Vishnubhotla2023, michel-etal-2024-improving}.
The current state-of-the-art on English novels is a recent reimplementation of BookNLP+ that uses SpanBERT \cite{Joshi2019} as the base encoder \cite{michel-etal-2024-improving}.

\begin{figure}
    \small
    \begin{enumerate}
\item[] \textcolor{olive}{"As soon as ever Mr. Bingley comes, my dear,"}  \underline{said Mrs. Bennet,} \textcolor{olive}{"you will wait on him of course."}
\item[] \textcolor{purple}{"No, no. You forced me into visiting him last year, and promised if I went to see him, he should marry one of my daughters..."}
\item[]  His wife represented to him how absolutely necessary such an attention would be from all the neighbouring gentlemen, on his returning to Netherfield.
\item[]  \textcolor{brown}{"'Tis an etiquette I despise,"} \underline{said he.}
    \end{enumerate}
    \caption{Excerpt of \textit{Pride and Prejudice} by Jane Austen (1813). Quotations are colored by quote type: \textcolor{olive}{explict}, \textcolor{purple}{implicit} and \textcolor{brown}{anaphoric}. Speaker information given by the narrator are underlined. Figure taken from \citet{michel-etal-2024-improving}.}
    \label{fig:quote_types}
\end{figure}

\paragraph{Memorization}
The zero-shot and few-shot performance of LLMs has often been attributed to memorization \cite{lee-etal-2022-deduplicating, Razeghi2022, Carlini2023}.
This raises important concerns in literary studies as some novels are present more often in the pretraining data of LLMs than others, creating discrepancies in downstream tasks \cite{chang2023}.
Assessing the impact of memorization on downstream tasks gives insights into LLMs capacity to generalize to unseen data, and is thus of critical importance.
\paragraph{Annotation Contamination}
Annotation contamination \cite{magar-schwartz-2022-data} occurs when downstream task \textit{evaluation data} (i.e. the exact annotations) is part of the LLMs pretraining corpus.
Methods such as Membership Inference Attacks \cite{Yeom2018, mireshghallah-etal-2022-quantifying, Shi2024} have been designed to evaluate an LLM ability to generate such data instances. 
This causes severe issues for security and privacy \cite{Carlini2021}, but also raises questions about zero-shot performance  \cite{Li2023a}.   
    

\begin{table*}[h!]
    \centering
    \small
    \begin{tabular}{l|ccc|ccc|ccc}
    \toprule
    & \multicolumn{3}{c}{$\textbf{PDNC}_1$} & \multicolumn{3}{c}{$\textbf{PDNC}_2$} & \multicolumn{3}{c}{$\textbf{Unseen}$} \\
    \midrule
    &  \textbf{All} & \textbf{Explicit} & \textbf{Other} &  \textbf{All} & \textbf{Explicit} & \textbf{Other} &  \textbf{All} & \textbf{Explicit} & \textbf{Other}\\
    \midrule
    ChatGPT & $71^{+}$ & - & $70^{+}$ & - & -  & - & -  & - & - \\ 
    BookNLP+ & $78.5$ \scriptsize{(4.0)} & $98.6$ \scriptsize{(1.6)} &  $68.9$ \scriptsize{(4.4)} & $79.2$ \scriptsize{(10.7)} & $93.3$ \scriptsize{(5.7)} & $69.6$ \scriptsize{(10.2)}  & $98.5$ & $99.1$ & $98.3$\\

    \midrule
    Llama-3 8b & $90.6$ \scriptsize{(5.2)} & $94.7$ \scriptsize{(2.9)} & $89.1$ \scriptsize{(5.7)} & $88.5$ \scriptsize{(4.0)} & $92.8$ \scriptsize{(2.1)} & $85.7$ \scriptsize{(4.9)} & $97.9$ & $97.5$ & $98.4$ \\
    
    \bottomrule
    \end{tabular}
    \caption{Quotation Attribution accuracy averaged over novels (standard deviations in parentheses) for Llama-3. We take the reported results from \citet{Su2023} for ChatGPT, and from \citet{michel-etal-2024-improving} for BookNLP+}
    \label{tab:qa_res}
\end{table*}

\section{Data}

We use the Project Dialogism Novel Corpus (PDNC) \cite{Vishnubhotla2022}, which contains 28 novels published between the 19th and 20th century, resulting in 37,131 quotes annotated manually with quotation attribution. PDNC is currently the largest dataset of quotation attribution.

PDNC quotes are categorized into three types: \textit{anaphoric} quotes, introduced with a speech verb and a pronoun or common noun, \textit{implicit} quotes, where no narrative details about the speaker are provided and \textit{explicit} quotes, which occur when the narrator identifies the speaker using a speech verb and a proper named-mention.
Examples are given in Figure~\ref{fig:quote_types}.


Among PDNC novels, 22 novels were released in July 2022 ($\text{PDNC}_1$), while 6 novels were added in June 2023 ($\text{PDNC}_2$).
The latter subset will be crucial to test for annotation contamination since it was released after Llama-3 8b's knowledge cutoff (March 2023).
Additionnaly, we fully annotated a new novel that was published after this cutoff. Following PDNC guidelines, one author annotated all quotes and a second author a subset of 5 chapters.
The inter-annotator agreement, measured by Cohen's $\kappa$ score, reached 97\% indicating almost perfect agreement.
A total of 1530 quotes were annotated.
We use this recent novel to assess Llama-3's generalization ability.



\section{Quotation Attribution}

We divide each novel by chapters, and chunk each chapter using $4096$ tokens with a stride of $1024$ tokens.
We modify the raw text by assigning a unique identifier to each quote starting from $1$ to $n$, where $n$ is the number of quotes in the chunk.
We also build a character-to-alias list using the gold character-list from PDNC that we include in the prompt.
Given the modified text and the list of character aliases, we prompt the model to predict the speaker of quotes $1,\dots,n$ sequentially.
We use \textit{Llama-3 8b Instruct} for all experiments, and test the 70b version on the Unseen novel as its annotations are not included in the larger model pretraining data.
More details are provided in Appendix~\ref{sec:appendix_d}.

\paragraph{Baselines} We compare to \citet{Su2023} ChatGPT's (\textit{gpt-3.5-turbo-0613}) Chain-of-Thought prompting strategy where the model is prompted with a target quote and its surrounding context.
We also compare to the current state-of-the-art on PDNC \cite{michel-etal-2024-improving}.
We use the official code to train BookNLP+ with the first cross-validation split of $\text{PDNC}_1$ that we further employ to attribute quotes in $\text{PDNC}_2$ and the unseen novel.

\begin{table*}[ht!]
    \centering
    \small
    \begin{tabular}{l|cc|cc|cc}
    \toprule
    & \multicolumn{2}{c}{\textbf{Accuracy (All)}} & \multicolumn{2}{c}{\textbf{Accuracy (Explicit)}} & \multicolumn{2}{c}{\textbf{Accuracy (Others)}} \\
    \midrule
    &  $\rho$ &  $(\texttt{Top}_5 - \texttt{Bot}_5)$ &  $\rho$ &   $(\texttt{Top}_5 - \texttt{Bot}_5)$ &  $\rho$ &  $(\texttt{Top}_5 - \texttt{Bot}_5)$ \\
    \midrule
    Name-Cloze & $0.15^{\phantom{\star}}$ & \xmark & $0.27^{\star}$ &  \xmark & $0.01^{\phantom{\star}}$ &  \xmark \\
    CSG-Memorization & $0.09^{\phantom{\star}}$ &  \xmark & $0.34^{\star}$ & \xmark  & $0.01^{\phantom{\star}}$ &  \xmark \\
    CSG-Reasoning & $0.52^{\star}$ & \checkmark & $0.21^{\phantom{\star}}$ &  \xmark & $0.43^{\star}$ & \xmark \\
    \bottomrule
    \end{tabular}
    \caption{Correlations (Spearman $\rho$) between quotation attribution accuracy and measures of memorization ($^\star$ indicates $p <0.05$), and statistical significance at $5\%$ from a Student t-test when testing for difference in expected attribution accuracies between top 5 most memorized books and bottom 5 least memorized books $(\texttt{Top}_5 - \texttt{Bot}_5)$.}
    \label{tab:corr_qa}
\end{table*}

\paragraph{Evaluation} 
We follow previous works \cite{Vishnubhotla2023, Su2023, michel-etal-2024-improving}, and focus on \textit{major} and \textit{intermediate} characters, which are characters that utter at least 10 quotes in a novel.
We present attribution accuracy on \textit{explicit} and \textit{other} quotes, (including both \textit{anaphoric} and \textit{implicit} utterances) \cite{Muzny2017, Vishnubhotla2022}.
Explicit utterances occur when the narrator indicates the speaker of a quote with a speech verb and a named mention, while anaphoric quotes are introduced with a speech verb and a pronoun or common noun.
When no narrative information is given about the speaker of the quote, we refer to those as implicit quotes.


\paragraph{Results} Table~\ref{tab:qa_res} shows surprisingly high performance for
Llama3-8b, increasing the overall attribution accuracy by up to $19$ points against ChatGPT on $\text{PDNC}_1$ and $12$ points against BookNLP+.
This gain is due to the large performance increase when attributing non-explicit quotes, that we also see on $\text{PDNC}_2$.
This suggests that Llama-3 might be able to solve complex cases of reasoning such as coreference resolution in a small context, or understanding discussion patterns.


On the Unseen novel, BookNLP+ performs slightly better than Llama-3 8b overall.
When increasing the model size to 70b, the performance increases to an almost perfect accuracy, and we identified only 3 wrong predictions out of 1442 quotes (note that we only consider \textit{major} and \textit{interemediate} characters).
The larger model appears to have improved reasoning abilities, yielding better attribution. While Llama-3 shows surprising performance on both subsets of PDNC, we question if those results are due to its reasoning abilities. 
Thus, we analyze the impact of memorization, reasoning and annotation contamination in the next section.




\section{The Impact of Memorization} 
\label{sec:memorization}

The extent to which LLMs have encountered books and annotations in their training data may influence and bias their assessment on downstream tasks \cite{razeghi-etal-2022-impact, chang2023, Li2023a}.
We thus carry out an evaluation of book memorization and annotation contamination.



\paragraph{Book Memorization.}
We use name-cloze accuracy \cite{chang2023} to quantify book memorization.
This methods prompts an LLM to identify a masked character name in a small passage of text.
Llama-3 8b achieves a 4\% average accuracy on PDNC, with 13 novels showing null accuracies.
Surprisingly, we found null name-cloze accuracies for canonical works such as 
\textit{The Picture of Dorian Gray}  compared to reported GPT-4 accuracies of $42$\%.
This questions name-cloze's validity for Llama-3 8b, leading us to propose a new metric: \textit{Corrupted-Speaker-Guessing} (CSG).


We design CSG as a speaker-guessing task, providing the model with the book's title, author, a passage, and a target quote.
We corrupt the passage by replacing the speaker’s name with a different gender-matching name that is not used in the book.
This pseudonymization approach has been used for example to build narrative-focused story embeddings \cite{hatzel-biemann-2024-story}.
When making a prediction, the LLM must decide whether to use contextual cues (\textit{reasoning}) or rely on memorized information to identify the correct speaker, despite the misleading contextual information.
More details and prompt examples are provided in Appendix~\ref{appendix:csg_details}


We validate CSG in two ways. First, we follow \citet{chang2023} and present the Spearman $\rho$ correlation between memorization metrics and the average number of search results for 10-grams randomly sampled from a book across Google, Bing, C4, and The Pile.
Significant correlations were found with all memorization measures (detailed in Appendix~\ref{appendix:corr_search}). Then, we ensured that all memorization metrics returned null accuracies on the unseen novel.


\paragraph{Impact on Quotation Attribution} We calculate Spearman $\rho$ correlations between quotation attribution accuracy and memorization and reasoning metrics.
We then identify the top 5 most and least memorized (or \textit{reasoned} in the case of CSG-Reasoning) books and test for differences in expected quotation attribution accuracy using a Student t-test.
Table~\ref{tab:corr_qa} shows positive correlations between memorization metrics and accuracy for explicit quotes, but not over all quotes.
These results suggest that book memorization does not explain Llama-3's impressive performance at attributing utterances of direct-speech, as also evidenced by high CSG-reasoning correlations. See Appendix~\ref{appendix:csg_per_novel} for detailed results per novel.

\paragraph{Annotation Contamination.}
We use \texttt{Min-K\%} \cite{Shi2024}, a popular contamination detection method, with 20\% randomly sampled annotation instances per novel.
For each data instance, we verbalize it in plain text, and then compute \texttt{Min-K\%} by averaging conditional probabilities of the \texttt{K}\% tokens with the lowest values in the sequence.

\begin{figure}
    \begin{small}
    \centering
    \includegraphics[width=1\linewidth]{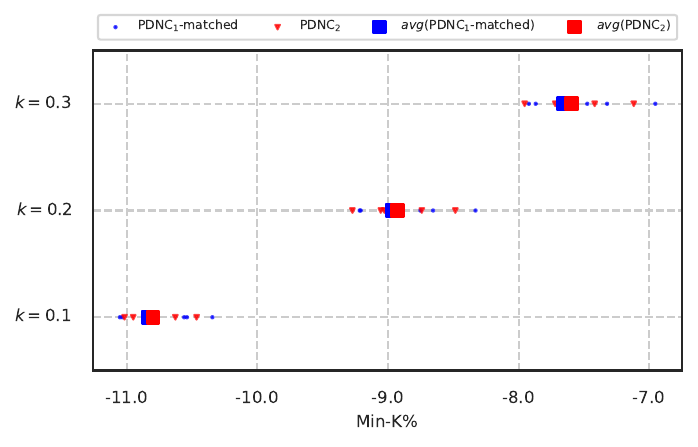}
    \caption{\texttt{Min-K\%} results for various values of \texttt{K} for $\text{PDNC}_2$ and each matched novel in $\text{PDNC}_1$.}
    \label{fig:matched}
    \end{small}
\end{figure}

A key challenge in analyzing Llama-3 probabilities is that annotation instances contain quotes and entities from novels, which can lead to variations in perplexity depending on the number of memorized passages from the book. 
To address this, we propose an econometrics-inspired approach: propensity score matching \cite{10.1093/biomet/70.1.41} to control the influence of book memorization when analyzing Llama-3 probabilities.
We begin by calculating a propensity score for each novel by fitting a logisitic regression, with the indicator of a novel being in $\text{PDNC}_2$ as the predictor. We include CSG-Memorization, name-cloze and \texttt{Min-K}\% as covariates, as well as overall quotation attribution accuracy, which may vary based on whether the annotations are memorized or not.
Predicted propensity scores reflect the likelihood of a novel belonging to $\text{PDNC}_2$, and hence indicate the probability that its annotations are unseen by Llama-3, given its degree of memorization.
For each novel in $\text{PDNC}_2$, we match a novel in $\text{PDNC}_1$ with the closest propensity score.
Figure~\ref{fig:matched} displays the average log-probabilities for each $\text{PDNC}_2$ novel and their $\text{PDNC}_1$ match.
We test for differences in expected value between the \texttt{Min-K}\% values with a paired Student t-test, and found no significant differences, suggesting that Llama-3 8b is unlikely to have memorized annotation instances of $\text{PDNC}_1$ (see Appendix~\ref{sec:appendix_c} for a detailed analysis).

\section{Conclusion}

We systematically evaluate Llama-3's zero-shot performance in quotation attribution, demonstrating that a simple Chain-Of-Thought approach accurately attributes direct-speech utterances from book chapters and significantly surpasses previous state-of-the-art models by a large margin.
Then, we analyze the reasons behind such performance by evaluating the impact of memorization on the downstream task.
Our results suggest that neither book memorization nor annotation contamination are key factors contributing to this improvement, suggesting Llama-3 as the current best system for quotation attribution in English literature.


\section{Limitations}

We proposed a new, task-specific and model-specific measure of book memorization.
While this measure shows a better capacity to recognize memorization than name-cloze accuracy when used with Llama-3 8b, we note that it is specific to literary texts, and that it suffers from one of the common downsides of this kind of measures: we can not be sure that instances of data have not been seen during pretraining.
Some novels in our corpus exhibit non-memorization, while we know that they are part of large corpus such as The Pile or C4, indicating that we could design better tests for book memorization.
Overall, we believe that the better way to test generalization of LLMs on a downstream task is to provide it with completely unseen data, which we tested by evaluating Llama-3 on a new, recently published novel.

Our metric, CSG, also labels prediction as a \textit{reasoning} class.
In reality, we can not be sure that the LLM is indeed \textit{reasoning} as a human would do, and we instead use this specific word to indicate that the LLM is processing contextual information, and is able to prioritize this contextual information over the uncorrupted passage it has memorized.
Besides, it is hard to understand why it prioritizes \textit{reasoning} over \textit{memorization}, and it is possible that larger models would prioritize more memorization. 

The significant improvement of Llama-3 over baselines such as BookNLP+ on quotation attribution creates new possibilities to better analyze large corpora of literary texts.
However, this improvement comes with longer inference times, taking up to a GPU hour for a single novel and limiting its impact for the study of massive corpora such as Project Gutenberg.
In comparison, BookNLP+ makes predictions in a few minutes for a novel.

In this work, we prompted Llama-3 with a predefined gold character-to-alias list. In real-world scenarios, this list is unlikely to be available.
Although approaches to build an alias list have been widely explored in the literature, our work does not mirror the full workflow of character discovery followed by quotation attribution.
\bibliography{anthology, custom, bilbio}

\begin{thebibliography}{36}
\providecommand{\natexlab}[1]{#1}

\bibitem[{Bamman et~al.(2024)Bamman, Chang, Lucy, and Zhou}]{bamman2024classificationlargelanguagemodels}
David Bamman, Kent~K. Chang, Li~Lucy, and Naitian Zhou. 2024.
\newblock \href {https://arxiv.org/abs/2410.12029} {On classification with large language models in cultural analytics}.
\newblock \emph{Preprint}, arXiv:2410.12029.

\bibitem[{Carlini et~al.(2023)Carlini, Ippolito, Jagielski, Lee, Tramer, and Zhang}]{Carlini2023}
Nicholas Carlini, Daphne Ippolito, Matthew Jagielski, Katherine Lee, Florian Tramer, and Chiyuan Zhang. 2023.
\newblock \href {https://arxiv.org/abs/2202.07646} {Quantifying memorization across neural language models}.
\newblock \emph{Preprint}, arXiv:2202.07646.

\bibitem[{Carlini et~al.(2021)Carlini, Tramer, Wallace, Jagielski, Herbert-Voss, Lee, Roberts, Brown, Song, Erlingsson, Oprea, and Raffel}]{Carlini2021}
Nicholas Carlini, Florian Tramer, Eric Wallace, Matthew Jagielski, Ariel Herbert-Voss, Katherine Lee, Adam Roberts, Tom Brown, Dawn Song, Ulfar Erlingsson, Alina Oprea, and Colin Raffel. 2021.
\newblock \href {https://arxiv.org/abs/2012.07805} {Extracting training data from large language models}.
\newblock \emph{Preprint}, arXiv:2012.07805.

\bibitem[{Chang et~al.(2023)Chang, Cramer, Soni, and Bamman}]{chang2023}
Kent Chang, Mackenzie Cramer, Sandeep Soni, and David Bamman. 2023.
\newblock \href {https://doi.org/10.18653/v1/2023.emnlp-main.453} {Speak, memory: An archaeology of books known to {C}hat{GPT}/{GPT}-4}.
\newblock In \emph{Proceedings of the 2023 Conference on Empirical Methods in Natural Language Processing}, pages 7312--7327, Singapore. Association for Computational Linguistics.

\bibitem[{Dubey et~al.(2024)Dubey, Jauhri, Pandey, Kadian, Al-Dahle, Letman, Mathur, Schelten, Yang, Fan et~al.}]{dubey2024llama}
Abhimanyu Dubey, Abhinav Jauhri, Abhinav Pandey, Abhishek Kadian, Ahmad Al-Dahle, Aiesha Letman, Akhil Mathur, Alan Schelten, Amy Yang, Angela Fan, et~al. 2024.
\newblock The llama 3 herd of models.
\newblock \emph{arXiv preprint arXiv:2407.21783}.

\bibitem[{Elson et~al.(2010)Elson, Dames, and McKeown}]{Elson2010}
David Elson, Nicholas Dames, and Kathleen McKeown. 2010.
\newblock \href {https://aclanthology.org/P10-1015} {Extracting social networks from literary fiction}.
\newblock In \emph{Proceedings of the 48th Annual Meeting of the Association for Computational Linguistics}, pages 138--147, Uppsala, Sweden. Association for Computational Linguistics.

\bibitem[{Hatzel and Biemann(2024)}]{hatzel-biemann-2024-story}
Hans~Ole Hatzel and Chris Biemann. 2024.
\newblock \href {https://doi.org/10.18653/v1/2024.emnlp-main.339} {Story embeddings {---} narrative-focused representations of fictional stories}.
\newblock In \emph{Proceedings of the 2024 Conference on Empirical Methods in Natural Language Processing}, pages 5931--5943, Miami, Florida, USA. Association for Computational Linguistics.

\bibitem[{Hobson et~al.(2024)Hobson, Zhou, Ruths, and Piper}]{hobson-etal-2024-story}
David~G Hobson, Haiqi Zhou, Derek Ruths, and Andrew Piper. 2024.
\newblock \href {https://doi.org/10.18653/v1/2024.emnlp-main.723} {Story morals: Surfacing value-driven narrative schemas using large language models}.
\newblock In \emph{Proceedings of the 2024 Conference on Empirical Methods in Natural Language Processing}, pages 12998--13032, Miami, Florida, USA. Association for Computational Linguistics.

\bibitem[{Joshi et~al.(2019)Joshi, Levy, Zettlemoyer, and Weld}]{Joshi2019}
Mandar Joshi, Omer Levy, Luke Zettlemoyer, and Daniel Weld. 2019.
\newblock \href {https://doi.org/10.18653/v1/D19-1588} {{BERT} for coreference resolution: Baselines and analysis}.
\newblock In \emph{Proceedings of the 2019 Conference on Empirical Methods in Natural Language Processing and the 9th International Joint Conference on Natural Language Processing (EMNLP-IJCNLP)}, pages 5803--5808, Hong Kong, China. Association for Computational Linguistics.

\bibitem[{Labatut and Bost(2019)}]{Labatut2019}
Vincent Labatut and Xavier Bost. 2019.
\newblock \href {https://doi.org/10.1145/3344548} {Extraction and analysis of fictional character networks: A survey}.
\newblock \emph{ACM Computing Surveys}, 52(5):1–40.

\bibitem[{Lee et~al.(2022)Lee, Ippolito, Nystrom, Zhang, Eck, Callison-Burch, and Carlini}]{lee-etal-2022-deduplicating}
Katherine Lee, Daphne Ippolito, Andrew Nystrom, Chiyuan Zhang, Douglas Eck, Chris Callison-Burch, and Nicholas Carlini. 2022.
\newblock \href {https://doi.org/10.18653/v1/2022.acl-long.577} {Deduplicating training data makes language models better}.
\newblock In \emph{Proceedings of the 60th Annual Meeting of the Association for Computational Linguistics (Volume 1: Long Papers)}, pages 8424--8445, Dublin, Ireland. Association for Computational Linguistics.

\bibitem[{Li and Flanigan(2023)}]{Li2023a}
Changmao Li and Jeffrey Flanigan. 2023.
\newblock \href {https://arxiv.org/abs/2312.16337} {Task contamination: Language models may not be few-shot anymore}.
\newblock \emph{Preprint}, arXiv:2312.16337.

\bibitem[{Magar and Schwartz(2022)}]{magar-schwartz-2022-data}
Inbal Magar and Roy Schwartz. 2022.
\newblock \href {https://doi.org/10.18653/v1/2022.acl-short.18} {Data contamination: From memorization to exploitation}.
\newblock In \emph{Proceedings of the 60th Annual Meeting of the Association for Computational Linguistics (Volume 2: Short Papers)}, pages 157--165, Dublin, Ireland. Association for Computational Linguistics.

\bibitem[{Michel et~al.(2024)Michel, Epure, Hennequin, and Cerisara}]{michel-etal-2024-improving}
Gaspard Michel, Elena~V. Epure, Romain Hennequin, and Christophe Cerisara. 2024.
\newblock \href {https://doi.org/10.18653/v1/2024.findings-emnlp.744} {Improving quotation attribution with fictional character embeddings}.
\newblock In \emph{Findings of the Association for Computational Linguistics: EMNLP 2024}, pages 12723--12735, Miami, Florida, USA. Association for Computational Linguistics.

\bibitem[{Mireshghallah et~al.(2022)Mireshghallah, Goyal, Uniyal, Berg-Kirkpatrick, and Shokri}]{mireshghallah-etal-2022-quantifying}
Fatemehsadat Mireshghallah, Kartik Goyal, Archit Uniyal, Taylor Berg-Kirkpatrick, and Reza Shokri. 2022.
\newblock \href {https://doi.org/10.18653/v1/2022.emnlp-main.570} {Quantifying privacy risks of masked language models using membership inference attacks}.
\newblock In \emph{Proceedings of the 2022 Conference on Empirical Methods in Natural Language Processing}, pages 8332--8347, Abu Dhabi, United Arab Emirates. Association for Computational Linguistics.

\bibitem[{Muzny et~al.(2017{\natexlab{a}})Muzny, Algee-Hewitt, and Jurafsky}]{Muzny2017a}
Grace Muzny, Mark Algee-Hewitt, and Dan Jurafsky. 2017{\natexlab{a}}.
\newblock \href {https://doi.org/10.1093/llc/fqx031} {{Dialogism in the novel: A computational model of the dialogic nature of narration and quotations}}.
\newblock \emph{Digital Scholarship in the Humanities}, 32:ii31--ii52.

\bibitem[{Muzny et~al.(2017{\natexlab{b}})Muzny, Fang, Chang, and Jurafsky}]{Muzny2017}
Grace Muzny, Michael Fang, Angel Chang, and Dan Jurafsky. 2017{\natexlab{b}}.
\newblock \href {https://aclanthology.org/E17-1044} {A two-stage sieve approach for quote attribution}.
\newblock In \emph{Proceedings of the 15th Conference of the {E}uropean Chapter of the Association for Computational Linguistics: Volume 1, Long Papers}, pages 460--470, Valencia, Spain. Association for Computational Linguistics.

\bibitem[{O{'}Keefe et~al.(2012)O{'}Keefe, Pareti, Curran, Koprinska, and Honnibal}]{okeefe-etal-2012-sequence}
Timothy O{'}Keefe, Silvia Pareti, James~R. Curran, Irena Koprinska, and Matthew Honnibal. 2012.
\newblock \href {https://aclanthology.org/D12-1072} {A sequence labelling approach to quote attribution}.
\newblock In \emph{Proceedings of the 2012 Joint Conference on Empirical Methods in Natural Language Processing and Computational Natural Language Learning}, pages 790--799, Jeju Island, Korea. Association for Computational Linguistics.

\bibitem[{Piper and Bagga(2024)}]{piper-bagga-2024-using}
Andrew Piper and Sunyam Bagga. 2024.
\newblock \href {https://doi.org/10.18653/v1/2024.wnu-1.4} {Using large language models for understanding narrative discourse}.
\newblock In \emph{Proceedings of the The 6th Workshop on Narrative Understanding}, pages 37--46, Miami, Florida, USA. Association for Computational Linguistics.

\bibitem[{Razeghi et~al.(2022{\natexlab{a}})Razeghi, Logan~IV, Gardner, and Singh}]{Razeghi2022}
Yasaman Razeghi, Robert~L Logan~IV, Matt Gardner, and Sameer Singh. 2022{\natexlab{a}}.
\newblock \href {https://doi.org/10.18653/v1/2022.findings-emnlp.59} {Impact of pretraining term frequencies on few-shot numerical reasoning}.
\newblock In \emph{Findings of the Association for Computational Linguistics: EMNLP 2022}, pages 840--854, Abu Dhabi, United Arab Emirates. Association for Computational Linguistics.

\bibitem[{Razeghi et~al.(2022{\natexlab{b}})Razeghi, Logan~IV, Gardner, and Singh}]{razeghi-etal-2022-impact}
Yasaman Razeghi, Robert~L Logan~IV, Matt Gardner, and Sameer Singh. 2022{\natexlab{b}}.
\newblock \href {https://doi.org/10.18653/v1/2022.findings-emnlp.59} {Impact of pretraining term frequencies on few-shot numerical reasoning}.
\newblock In \emph{Findings of the Association for Computational Linguistics: EMNLP 2022}, pages 840--854, Abu Dhabi, United Arab Emirates. Association for Computational Linguistics.

\bibitem[{Rosenbaum and Rubin(1983)}]{10.1093/biomet/70.1.41}
Paul~R. Rosenbaum and Donald~B. Rubin. 1983.
\newblock \href {https://doi.org/10.1093/biomet/70.1.41} {{The central role of the propensity score in observational studies for causal effects}}.
\newblock \emph{Biometrika}, 70(1):41--55.

\bibitem[{Shi et~al.(2024)Shi, Ajith, Xia, Huang, Liu, Blevins, Chen, and Zettlemoyer}]{Shi2024}
Weijia Shi, Anirudh Ajith, Mengzhou Xia, Yangsibo Huang, Daogao Liu, Terra Blevins, Danqi Chen, and Luke Zettlemoyer. 2024.
\newblock \href {https://arxiv.org/abs/2310.16789} {Detecting pretraining data from large language models}.
\newblock \emph{Preprint}, arXiv:2310.16789.

\bibitem[{Sims and Bamman(2020)}]{Sims2020}
Matthew Sims and David Bamman. 2020.
\newblock \href {https://doi.org/10.18653/v1/2020.emnlp-main.47} {Measuring information propagation in literary social networks}.
\newblock In \emph{Proceedings of the 2020 Conference on Empirical Methods in Natural Language Processing (EMNLP)}, pages 642--652, Online. Association for Computational Linguistics.

\bibitem[{Soni et~al.(2023)Soni, Sihra, Evans, Wilkens, and Bamman}]{Soni2023}
Sandeep Soni, Amanpreet Sihra, Elizabeth Evans, Matthew Wilkens, and David Bamman. 2023.
\newblock \href {https://doi.org/10.18653/v1/2023.acl-long.655} {Grounding characters and places in narrative text}.
\newblock In \emph{Proceedings of the 61st Annual Meeting of the Association for Computational Linguistics (Volume 1: Long Papers)}, pages 11723--11736, Toronto, Canada. Association for Computational Linguistics.

\bibitem[{Stammbach et~al.(2022)Stammbach, Antoniak, and Ash}]{Stammbach2022}
Dominik Stammbach, Maria Antoniak, and Elliott Ash. 2022.
\newblock \href {https://doi.org/10.18653/v1/2022.wnu-1.6} {Heroes, villains, and victims, and {GPT}-3: Automated extraction of character roles without training data}.
\newblock In \emph{Proceedings of the 4th Workshop of Narrative Understanding (WNU2022)}, pages 47--56, Seattle, United States. Association for Computational Linguistics.

\bibitem[{Su et~al.(2023)Su, Xu, Xu, Li, and Huangfu}]{Su2023}
Zhenlin Su, Liyan Xu, Jin Xu, Jiangnan Li, and Mingdu Huangfu. 2023.
\newblock \href {https://arxiv.org/abs/2312.14590} {Sig: Speaker identification in literature via prompt-based generation}.
\newblock \emph{Preprint}, arXiv:2312.14590.

\bibitem[{Underwood(2023)}]{Underwood2023}
Ted Underwood. 2023.
\newblock \href {https://tedunderwood.com/2023/03/19/using-gpt-4-to-measure-the-passage-of-time-in-fiction/} {Using gpt-4 to measure the passage of time in fiction}.

\bibitem[{Vishnubhotla et~al.(2022)Vishnubhotla, Hammond, and Hirst}]{Vishnubhotla2022}
Krishnapriya Vishnubhotla, Adam Hammond, and Graeme Hirst. 2022.
\newblock \href {https://aclanthology.org/2022.lrec-1.628} {The project dialogism novel corpus: A dataset for quotation attribution in literary texts}.
\newblock In \emph{Proceedings of the Thirteenth Language Resources and Evaluation Conference}, pages 5838--5848, Marseille, France. European Language Resources Association.

\bibitem[{Vishnubhotla et~al.(2023)Vishnubhotla, Rudzicz, Hirst, and Hammond}]{Vishnubhotla2023}
Krishnapriya Vishnubhotla, Frank Rudzicz, Graeme Hirst, and Adam Hammond. 2023.
\newblock \href {https://doi.org/10.18653/v1/2023.acl-short.64} {Improving automatic quotation attribution in literary novels}.
\newblock In \emph{Proceedings of the 61st Annual Meeting of the Association for Computational Linguistics (Volume 2: Short Papers)}, pages 737--746, Toronto, Canada. Association for Computational Linguistics.

\bibitem[{Wang et~al.(2024)Wang, Ning, Pan, Wu, Guo, Deng, Bao, Wang, and Zhang}]{Wang2024}
Cunxiang Wang, Ruoxi Ning, Boqi Pan, Tonghui Wu, Qipeng Guo, Cheng Deng, Guangsheng Bao, Qian Wang, and Yue Zhang. 2024.
\newblock \href {https://arxiv.org/abs/2403.12766} {Novelqa: A benchmark for long-range novel question answering}.
\newblock \emph{Preprint}, arXiv:2403.12766.

\bibitem[{Wei et~al.(2022)Wei, Wang, Schuurmans, Bosma, brian ichter, Xia, Chi, Le, and Zhou}]{wei2022chain}
Jason Wei, Xuezhi Wang, Dale Schuurmans, Maarten Bosma, brian ichter, Fei Xia, Ed~H. Chi, Quoc~V Le, and Denny Zhou. 2022.
\newblock \href {https://openreview.net/forum?id=_VjQlMeSB_J} {Chain of thought prompting elicits reasoning in large language models}.
\newblock In \emph{Advances in Neural Information Processing Systems}.

\bibitem[{Xu et~al.(2023)Xu, Ping, Wu, McAfee, Zhu, Liu, Subramanian, Bakhturina, Shoeybi, and Catanzaro}]{Xu2023}
Peng Xu, Wei Ping, Xianchao Wu, Lawrence McAfee, Chen Zhu, Zihan Liu, Sandeep Subramanian, Evelina Bakhturina, Mohammad Shoeybi, and Bryan Catanzaro. 2023.
\newblock Retrieval meets long context large language models.
\newblock \emph{arXiv preprint arXiv:2310.03025}.

\bibitem[{Yeom et~al.(2018)Yeom, Giacomelli, Fredrikson, and Jha}]{Yeom2018}
Samuel Yeom, Irene Giacomelli, Matt Fredrikson, and Somesh Jha. 2018.
\newblock \href {https://arxiv.org/abs/1709.01604} {Privacy risk in machine learning: Analyzing the connection to overfitting}.
\newblock \emph{Preprint}, arXiv:1709.01604.

\bibitem[{Yu et~al.(2023)Yu, Li, Yao, Pang, Zhou, Xiao, Meng, and Zhou}]{Yu2023a}
Mo~Yu, Jiangnan Li, Shunyu Yao, Wenjie Pang, Xiaochen Zhou, Zhou Xiao, Fandong Meng, and Jie Zhou. 2023.
\newblock \href {https://arxiv.org/abs/2305.10156} {Personality understanding of fictional characters during book reading}.
\newblock \emph{Preprint}, arXiv:2305.10156.

\bibitem[{Zhao et~al.(2024)Zhao, Zhang, Li, Zhu, Li, He, and Gui}]{Zhao2024}
Runcong Zhao, Wenjia Zhang, Jiazheng Li, Lixing Zhu, Yanran Li, Yulan He, and Lin Gui. 2024.
\newblock \href {https://aclanthology.org/2024.eacl-demo.10} {{N}arrative{P}lay: Interactive narrative understanding}.
\newblock In \emph{Proceedings of the 18th Conference of the European Chapter of the Association for Computational Linguistics: System Demonstrations}, pages 82--93, St. Julians, Malta. Association for Computational Linguistics.

\end{thebibliography}


\appendix
\newpage
\section{Method Details - Quotation Attribution}
\label{sec:appendix_d}

We divide novels in chapters, and build chunks of text of length $4096$ tokens with a stride of $1024$ tokens.
If an entire chapter is less than $4096$ tokens, then we use all tokens in this chapter and do not use striding for the next chunk.
That is we only use striding when chapters are longer than $4096$ tokens.
All quotes in a chunk need to be predicted by the model.

With the above chunk construction, some quotes will be predicted twice when striding is used.
We experiment with two approaches:
\begin{itemize}
    \item[1.] We consider only the first prediction of a quote, i.e. the first time it appears in a chunk.
    \item[2.] We propose an incremental prompting strategy, where predictions of overlapping quotes are also given as contextual information, and we prompt the LLM to predict all quotes in a chunk, refining its prediction if necessary.
\end{itemize}

In all cases, we use Chain-of-Thought prompting, and prompt the model with the gold character-to-alias list.
We tested without using this list, but we realized that the model was often predicting aliases that were not in this list, which made the attribution to a character ID a lot harder.
We found that using the gold character-to-alias list is the most straightforward way to restrict the generation to a candidate name, but also makes our results an upper-bound when evaluating the end-to-end workflow of quotation attribution that also includes building a silver character-to-alias list.
Note that the gold character list is also used by other baselines (ChatGPT and BookNLP+), making the comparison with our approach still fair.

A prompt example used in strategy (1.) is displayed in Figure~\ref{fig:first_prompt} and an incremental prompt example used when there are overlapping quotes in strategy (2.) is displayed in Figure~\ref{fig:incr_prompt}.

The model output is a JSON string, with unique quote identifiers as keys and predicted names as values.
In particular, we use the character-to-alias list to replace the predicted name with their canonical character ID (which is our gold label).
If the model generates a name that is not an alias, we consider its predictions as wrong (\textit{i.e.} we do not use any lenient metrics such as substring matching). 

Results for both strategies on PDNC$_2$ are displayed in Table~\ref{tab:ablation}.
We found that the incremental strategy led to slightly better results on this subset of novels, and thus used it for all experiments.

\begin{table}[h!]
    \centering
    \begin{tabular}{l|ccc}
        \toprule
         &  \textbf{All} &  \textbf{Explicit}  &  \textbf{Others}  \\
         \midrule
         Strategy 1. & $87.6$ \scriptsize(3.9) & $92.0$ \scriptsize(2.5) & $84.7$ \scriptsize(4.9)\\ 
         Strategy 2. & $88.5$ \scriptsize(4.0) & $92.8$ \scriptsize(2.1) & $85.7$ \scriptsize(4.9)\\
         \bottomrule
    \end{tabular}
    \caption{Average Quotation Attribution accuracy on $\text{PDNC}_2$, with (standard deviation) for both strategies.}
    \label{tab:ablation}
\end{table}


\section{Method Details - CSG}
\label{appendix:csg_details}

We designed \textit{Corrupted-Speaker-Guessing} by finding out the really low/null name-cloze accuracies of Llama-3 8b on PDNC.
These results suggests that Llama-3 has not \textit{exactly memorized} some canonical PDNC novels.
To avoid a similar situation where CSG returns null accuracies, we also provide book-level metadata as contextual information to be able to catch \textit{weaker memorization}.
CSG prompts an LLM with a corrupted passage of a book, the book's title and author, and a target quote appearing in the passage.
The passage contains 10 sentences before and after the target quote (we use SpaCy to segment sentences).
It tasks the LLM to find the speaker of the target quote.
To corrupt the original passage, we apply the following modifications: 

\begin{itemize}
    \item[1.] We find all proper named mentions of the speaker, using the gold character-to-alias list.
    \item[2.] We replace all proper named mentions of the speaker with another name, matching its gender. We use two first names for each gender: ``Henry'' or ``Joseph'' and ``Emma'' or ``Elizabeth''. We also use three last names: ``Stone'', ``Walker'' and ``Smith''.
    We use combinations of first and last names such that none of these names appear in the novel.
    Finally, we kept all honorifics when replacing (``Miss Bates'' $\xrightarrow[]{}$ ``Miss Smith'').
\end{itemize}

Note that this process was done manually by one of the author and that we never used ``Emma Stone'' or other celebrity names that are likely to appear more frequently on the web.

We use two different prompts, depending on whether the target quote is an explicit quote or non-explicit.
In the case of explicit quotes, we formulate the task as a cloze, replacing all named mentions and masking the referring expression (``said [MASK]'').
An example is provided in Figure~\ref{fig:cloze_prompt}.
For other quote types, we do not use masking and use the prompt provided in Figure~\ref{fig:imp_prompt} and Figure~\ref{fig:ana_prompt}.
We ensure that there is at least one named mention of the speaker in the corrupted passage, such that contextual information should point to the corrupted character name as the speaker.

For each quote type (explicit, anaphoric and implicit), we randomly sample 100 quotes and their associated corrupted passages, and prompt the model to find the speaker of the target quote.
Given the model's prediction, we calculate two types of accuracy:
\begin{itemize}
    \item Memorization accuracy: when the model predicts the true speaker name, even though the passage does not contain any named mention of this speaker.
    \item Reasoning accuracy: when the model uses contextual information to predict the corrupted speaker name.
\end{itemize} 

We calculate CSG-Memorization and CSG-Reasoning accuracies by averaging each accuracy over all quote types.

\section{CSG Validation}
\label{appendix:corr_search}

One of the validation of CSG was done following \cite{chang2023}, by evaluating the correlation between (a proxy of) the frequency of of a novel on the web and its memorization accuracy.
We present in Table~\ref{tab:corr_search} all correlation results between the average number of search results of random $10$-grams on different databases, and memorization metrics.
We do not have access to the custom search APIs that were used in \citet{chang2023}, so we instead directly use their reported number of searches for each endpoint.
We gathered data for a subset of 16 PDNC novels that were also used by \cite{chang2023}, and calculate Spearman $\rho$ correlations between the memorization measures and the average number of search results.

\begin{table}[t!]
    \centering
    \small
    
    \begin{tabular}{l|cccc}
    \toprule
         &  Google & Bing & C4 & Pile \\
         \midrule
    Name-Cloze & 0.42 & 0.55 & 0.75 & 0.57\\ 
    CSG-Mem & 0.54 & 0.3 & 0.42 & 0.61 \\
    $\;\;\;\;$ \texttt{Cloze Only} & 0.65 & 0.44 & 0.45 & 0.53 \\
    \bottomrule
    \end{tabular}
    \caption{Correlation (Spearman $\rho$) between Llama-3 memorization measures and number of search results in Google, Bing, C4 and the Pile. All coefficients are significative except for CSG-Mem and Bing.}
    \label{tab:corr_search}
\end{table}

\section{Results per Novel for CSG and Name-Cloze}
\label{appendix:csg_per_novel}

We present in Table~\ref{tab:all_mem_res} all memorization and reasoning accuracies.
We also chose to display the CSG-Memorization accuracy with the cloze prompt (with explicit quotes) as it holds interesting properties: we found similar conclusions when replacing CSG-Memorization with the cloze variant of CSG-Memorization.
This cloze variant is more practical, as automatically finding speakers of explicit quotes in novels is usually the easiest attribution task among all quote types, as shown by all systems accuracy.
Therefore, one can use only CSG-Memorization Cloze as a measure of book memorization, removing the need for annotating all quote types to measure the full CSG-Memorization.

\section{Annotation Contamination Results per Novel}
\label{sec:appendix_c}

We calculate \texttt{Min-K}\% by verbalizing instances of data.
We present in Figure~\ref{fig:verbalized} an example of how we verbalize an instance of data.
We then calculate the conditional log-probabilities of each token in the verbalized sequence, and average the $k$\% lowest log-probabilities in the sequence, for $k=10, 20, 30$.

Given each novel in $\text{PDNC}_2$ and their  $\text{PDNC}_1$ match, we conduct a paired paired Student t-test and test for difference in expected \texttt{Min-K}\% values.
We found no statistical differences ($t=0.54$, $p=0.3$).

Other approaches to detect contamination involves a chronological analysis \cite{Li2023a}, comparing downstream performance on a set of data that is known to be inside the pretraining corpus to the performance on a set not included during pretraining.
We follow the same approach as described in the Annotation Contamination paragraph of Section~\ref{sec:memorization}, but instead define the outcome variable to be the quotation attribution accuracy rather than Min-K\% when matching with propensity score.
We found no significant differences in the expected values of quotation attribution accuracy ($t=0.75$, $p=0.25$) using a paired t-test from matched novels.

\section{Computing Information}
We used a 32-core Intel Xeon Gold 6244 CPU @ 3.60GHz CPU with 128GB RAM equipped with 3 RTX A5000 GPUs with 24GB RAM.
We used a single RTX A5000 for all \texttt{Llama3-8b} experiments.
We used the 8-bits version of \texttt{Llama3-8b-Instruct} using the \textit{BitsAndBytes} library.
The peak memory used was around 14GB of RAM.
We employ a relatively large contextual window, and ask the model to generates long attribution lists.
Thus, we observed quite large inference times, and processing entire novels varied from 10 minutes to an hour.
For the \texttt{Llama3-70b} experiments, we used one A100-80GB and used the 4-bits quantized version \texttt{Meta-Llama-3-70B-Instruct-Q4\_K\_M.gguf}.

\begin{figure}[h!]
    \small
\begin{framed}
You will be given a passage of the book Persuasion written by Jane Austen that you have seen in your training data. Find the proper name that fills the [MASK] token. This name is a proper name (not a pronoun or any other word). You must make a guess, even if you are uncertain. Do not explain your reasoning.
\newline 

You must format your answer in <speaker>[SPEAKER]<\\speaker> tags.
\newline 

Passage:
\newline

[$\dots$]
\newline

"It was my friend Mrs Rooke; Nurse Rooke; who, by-the-bye, had a great curiosity to see you, and was delighted to be in the way to let you in. She came away from Marlborough Buildings only on Sunday; and she it was who told me you were to marry Mr Elliot.  She had had it from Mrs Wallis herself, which did not seem bad authority.  She sat an hour with me on Monday evening, and gave me the whole history." \textbf{"The whole history," repeated [MASK]}, laughing.  "She could not make a very long history, I think, of one such little article of unfounded news."
\newline 

Mrs Smith said nothing.
\newline 

"But," \textbf{continued Emma}, presently, "though there is no truth in my having this claim on Mr Elliot, I should be extremely happy to be of use to you in any way that I could.  Shall I mention to him your being in Bath?  Shall I take any message?"
\newline 

[$\dots$]
\newline 

Target quote:

"The whole history,"
\end{framed}
    \caption{Example of a CSG prompt with an explicit quote. Here, the character \textit{Anne Elliot} from \textit{Persuasion} is replaced by \textit{Emma}.}
    \label{fig:cloze_prompt}
\end{figure}

\begin{figure}[t!]
    \small
\begin{framed}
\textbf{Raw Data:} "Q0","and what is the use of a book, without pictures or conversations?","[\'and what is the use of a book,\', \'without pictures or conversations?\']","[[254, 284], [301, 335]]","Alice","[]","Explicit","thought Alice","[[], []]","[[], []]","[[], []]"

\phantom{text}

\textbf{Verbalized Data:} quoteID: Q0; quoteText: and what is the use of a book, without pictures or conversations?; subQuotationList: ['and what is the use of a book,', 'without pictures or conversations?']; quoteByteSpans: [[254, 284], [301, 335]]; speaker: Alice; addressees: []; quoteType: Explicit; referringExpression: thought Alice; mentionTextsList: [[], []]; mentionSpansList: [[], []]; mentionEntitiesList: [[], []]

\end{framed}
    \caption{Example of a verbalized instance of data.}
    \label{fig:verbalized}
\end{figure}

\begin{figure}[t!]
    \small
\begin{framed}
You will be given a passage of the book Persuasion written by Jane Austen that you have seen in your training data. Find the true speaker name of the target quote. This name is a proper name (not a pronoun or any other word). You must make a guess, even if you are uncertain. Do not explain your reasoning.
\newline 

You must format your answer in <speaker>[SPEAKER]<\\speaker> tags.
\newline 

Passage:
\newline 

[$\dots$]
\newline 

\textbf{Captain Stone} left his seat, and walked to the fire-place; probably for the sake of walking away from it soon afterwards, and taking a station, with less bare-faced design, by Anne.
\newline

"You have not been long enough in Bath," said he, "to enjoy the evening parties of the place."
\newline

"Oh! no.  The usual character of them has nothing for me.  I am no card-player."
\newline

"You were not formerly, I know.  You did not use to like cards; but time makes many changes."
\newline

"I am not yet so much changed," cried Anne, and stopped, fearing she hardly knew what misconstruction.  After waiting a few moments he said, and as if it were the result of immediate feeling, "It is a period, indeed!  Eight years and a half is a period."
\newline

[$\dots$]
\newline

Target quote:
\newline

"You were not formerly, I know.  You did not use to like cards; but time makes many changes."

\end{framed}
    \caption{Example of a CSG prompt with an implicit quote. Here, the character \textit{Captain Wentworth} from \textit{Persuasion} is replaced by \textit{Captain Stone}.}
    \label{fig:imp_prompt}
\end{figure}

\begin{figure}[t!]
    \small
\begin{framed}
You will be given a passage of the book Persuasion written by Jane Austen that you have seen in your training data. Find the true speaker name of the target quote. This name is a proper name (not a pronoun or any other word). You must make a guess, even if you are uncertain. Do not explain your reasoning.
\newline 

You must format your answer in <speaker>[SPEAKER]<\\speaker> tags.
\newline 

Passage:
\newline 

[$\dots$]
\newline 

Charles shewed himself at the window, all was ready, their visitor had bowed and was gone, the Miss Musgroves were gone too, suddenly resolving to walk to the end of the village with the sportsmen:  the room was cleared, and \textbf{Emma might finish her breakfast as she could}.
\newline 

"It is over! it is over!" \textbf{she repeated to herself} again and again, in nervous gratitude.  "The worst is over!"
\newline 

[$\dots$]
\newline

Target quote:
\newline

"The worst is over!"

\end{framed}
    \caption{Example of a CSG prompt with an anaphoric quote. Here, the character \textit{Anne Elliot} from \textit{Persuasion} is replaced by \textit{Emma}.}
    \label{fig:ana_prompt}
\end{figure}

\begin{table*}[h!]
\centering 
\footnotesize
\begin{tabular}{ll|ccc|c}
\toprule
Title & Author & Name-Cloze & CSG-M & CSG-M (Cloze) & CSG-R \\
\midrule
The Age of Innocence & Edith Wharton & 0.0 & 0.27 & 0.27 & 0.5 \\
Pride and Prejudice & Jane Austen & 0.1 & 0.23 & 0.27 & 0.59 \\
The Picture Of Dorian Gray & Oscar Wilde & 0.0 & 0.22 & 0.44 & 0.48 \\
The Awakening & Kate Chopin & 0.0 & 0.21 & 0.28 & 0.49 \\
Emma & Jane Austen & 0.19 & 0.2 & 0.24 & 0.55 \\
Daisy Miller & Henry James & 0.0 & 0.19 & 0.46 & 0.7 \\
A Room With A View & E. M. Forster & 0.0 & 0.17 & 0.24 & 0.53 \\
The Sun Also Rises & Ernest Hemingway & 0.01 & 0.17 & 0.34 & 0.5 \\
Sense and Sensibility & Jane Austen & 0.04 & 0.16 & 0.16 & 0.7 \\
Northanger Abbey & Jane Austen & 0.03 & 0.12 & 0.2 & 0.64 \\
Anne Of Green Gables & Lucy M. Montgomery & 0.02 & 0.12 & 0.3 & 0.75 \\
Alice’s Adventures in Wonderland & Lewis Carroll & 0.47 & 0.12 & 0.27 & 0.61 \\
Persuasion & Jane Austen & 0.0 & 0.11 & 0.21 & 0.62 \\
The Sign of the Four & Arthur Conan Doyle & 0.03 & 0.06 & 0.08 & 0.34 \\
The Invisible Man & Herbert George Wells & 0.02 & 0.06 & 0.16 & 0.88 \\
Howards End & Edward Morgan Forster & 0.0 & 0.05 & 0.09 & 0.53 \\
The Mysterious Affair At Styles & Agatha Christie & 0.0 & 0.03 & 0.06 & 0.63 \\
A Handful Of Dust & Evelyn Waugh & 0.0 & 0.02 & 0.0 & 0.57 \\
The Gambler & F. M. Dostoevsky & 0.01 & 0.02 & 0.04 & 0.58 \\
Night and Day & Virginia Woolf & 0.0 & 0.01 & 0.03 & 0.78 \\
The Man Who Was Thursday & Gilbert K.  Chesterton & 0.0 & 0.0 & 0.0 & 0.67 \\
The Sport of the Gods & Paul Laurence Dunbar & 0.0 & 0.0 & 0.0 & 0.64 \\
\midrule
A Passage to India & Edward Morgan Forster & 0.0 & 0.12 & 0.17 & 0.43 \\
Mansfield Park & Jane Austen & 0.0 & 0.09 & 0.13 & 0.59 \\
Winnie-The-Pooh & Alan Alexander Milne & 0.06 & 0.07 & 0.14 & 0.66 \\
Where Angels Fear to Tread & Edward Morgan Forster & 0.0 & 0.04 & 0.08 & 0.57 \\
Oliver Twist & Charles Dickens & 0.07 & 0.03 & 0.06 & 0.71 \\
Hard Times & Charles Dickens & 0.02 & 0.01 & 0.01 & 0.78 \\
\midrule
Dark Corners & Katie Rush & 0.0 & 0.0 & 0.0 & 0.84 \\
\bottomrule
\end{tabular}
\caption{All Memorization and Reasoning accuracies calculated with Llama-3 8b per novel. Top: PDNC$_1$, Middle: PDNC$_2$, Bottom: Unsenn novel.}
\label{tab:all_mem_res}
\end{table*}

\begin{figure*}[t!]
    \small
\begin{framed}
\textbf{Instruction:} You are an excellent linguist working in the field of literature. I will provide you with a passage of a book where some quotes have unique identifiers marked by headers '|quote\_id|'. Your are tasked to build a list of quote attributions by sequentially attributing the marked quotes to their speaker.
\newline

\textbf{Passage:}

---

Chapter 8
\newline

From this time Captain Wentworth and Anne Elliot were repeatedly in the same circle.  They were soon dining in company together at Mr Musgrove's, for the little boy's state could no longer supply his aunt with a pretence for absenting herself; and this was but the beginning of other dinings and other meetings.
\newline

Whether former feelings were to be renewed must be brought to the proof; former times must undoubtedly be brought to the recollection of each; they could not but be reverted to; the year of their engagement could not but be named by him, in the little narratives or descriptions which conversation called forth.  His profession qualified him, his disposition lead him, to talk; and |1|"That was in the year six;"|1| |2|"That happened before I went to sea in the year six,"|2| occurred in the course of the first evening they spent together: and though his voice did not falter, and though she had no reason to suppose his eye wandering towards her while he spoke, Anne felt the utter impossibility, from her knowledge of his mind, that he could be unvisited by remembrance any more than herself.  There must be the same immediate association of thought, though she was very far from conceiving it to be of equal pain.
\newline 

[$\dots$]
\newline

|50|"Aye, to be sure.  Yes, indeed, oh yes!  I am quite of your opinion, Mrs Croft,"|50| was Mrs Musgrove's hearty answer.  |51|"There is nothing so bad as a separation.  I am quite of your opinion.  I know what it is, for Mr Musgrove always attends the assizes, and I am so glad when they are over, and he is safe back again."|51|
\newline

The evening ended with dancing.  On its being proposed, Anne offered her services, as

---

\textbf{Step 1:} Attribute sequentially each quote to their speaker.

\textbf{Step 2:} Match each speaker found in the previous step with one of the following name:

\textbf{Names}

---

Admiral Croft=The Admiral=Admiral

Anne Elliot=Miss Anne=Miss Anne Elliot=Anne

Captain Harville=Harville

Captain Wentworth=Wentworth=Frederick Wentworth=Frederick

Charles Hayter=Hayter

Charles Musgrove

Elizabeth

Henrietta Musgrove=Henrietta

Lady Dalrymple=Dalrymple

Lady Russell=Russell

Louisa Musgrove=Louisa

Mary Musgrove=Mary

Mr Shepherd=Shepherd=John Shepherd

Mrs Clay=Clay=Penelope

Mrs Musgrove=Musgrove

Mrs Smith=Hamilton=Smith=Miss Hamilton

Sir Walter Elliot=Walter Elliot=Sir Walter=Walter

Sophia Croft=Sister Of Captian Wentworth=Croft=Mrs Croft

The Waiter=Waiter

William Walter Elliot=William=Mr 
Elliot=Elliot

---

\textbf{Step 3:} Replace the speakers found in Step 1 with their matching name found in Step 2. Your answer should follow this JSON format:

\{

'quote\_id\_1' : 'predicted\_speaker\_1',

'quote\_id\_2' : 'predicted\_speaker\_2'

\}

Your answer should only contain the output of \textbf{Step 3} and can only contain quote identifiers and speakers. Never generate quote content and don't explain your reasoning.

\end{framed}
    \caption{Example of a prompt used when there are no overlapping quotes. We also only use this prompt when experiment without incremental updating. The novel here is \textit{Persuasion}.}
    \label{fig:first_prompt}
\end{figure*}

\begin{figure*}[t!]
    \small
\begin{framed}
\textbf{Instruction:} You are an excellent linguist working in the field of literature. I will provide you with a passage of a book where some quotes have unique identifiers marked by headers '|quote\_id|'. You will also be provided a list of characters and their aliases, and previous predictions. Your are tasked to build a list of quote attributions by sequentially attributing the marked quotes to their speaker.
\newline

\textbf{Passage:}

---

|1|"then?"|1|
\newline

|2|"All merged in my friendship, Sophia.  I would assist any brother officer's wife that I could, and I would bring anything of Harville's from the world's end, if he wanted it.  But do not imagine that I did not feel it an evil in itself."|2|
\newline

|3|"Depend upon it, they were all perfectly comfortable."|3|
\newline

|4|"I might not like them the better for that perhaps.  Such a number of women and children have no right to be comfortable on board."|4|
\newline 

[$\dots$]
\newline 

|19|"I beg your pardon, madam, this is your seat;"|19| and though she immediately drew back with a decided negative, he was not to be induced to sit down again.
\newline

Anne did not wish for more of such looks and speeches.  His cold politeness, his ceremonious grace, were worse than anything.

---

\textbf{Previous predictions:}

---

\{ '2': 'pred\_0', '4':  'pred\_1', '6':  'pred\_2', '11':  'pred\_3', '12':  'pred\_4' \}

---

\textbf{Step 1:} Attribute sequentially each quote to their speaker. Update the previous predictions if you think it contains wrong speaker prediction.

\textbf{Step 2:} Match each speaker found in the previous step with one of the following name:

\textbf{Names}

---

Admiral Croft=The Admiral=Admiral

Anne Elliot=Miss Anne=Miss Anne Elliot=Anne

Captain Harville=Harville

Captain Wentworth=Wentworth=Frederick Wentworth=Frederick

Charles Hayter=Hayter

Charles Musgrove

Elizabeth

Henrietta Musgrove=Henrietta

Lady Dalrymple=Dalrymple

Lady Russell=Russell

Louisa Musgrove=Louisa

Mary Musgrove=Mary

Mr Shepherd=Shepherd=John Shepherd

Mrs Clay=Clay=Penelope

Mrs Musgrove=Musgrove

Mrs Smith=Hamilton=Smith=Miss Hamilton

Sir Walter Elliot=Walter Elliot=Sir Walter=Walter

Sophia Croft=Sister Of Captian Wentworth=Croft=Mrs Croft

The Waiter=Waiter

William Walter Elliot=William=Mr 
Elliot=Elliot

---

\textbf{Step 3:} Replace the speakers found in Step 1 with their matching name found in Step 2. Your answer should follow this JSON format:

\{

'quote\_id\_1' : 'predicted\_speaker\_1',

'quote\_id\_2' : 'predicted\_speaker\_2'

\}

Your answer should only contain the output of \textbf{Step 3} and can only contain quote identifiers and speakers. Never generate quote content and don't explain your reasoning.

\end{framed}
    \caption{Example of an incremental prompt used when there are overlapping quotes between the last chunk and the current chunk. The novel here is \textit{Persuasion}.}
    \label{fig:incr_prompt}
\end{figure*}

\end{document}